\pdfoutput=1

\documentclass[11pt]{article}

\usepackage{EMNLP2022}

\usepackage{times}
\usepackage{latexsym}

\usepackage[T1]{fontenc}

\usepackage[utf8]{inputenc}
\usepackage{microtype}
\usepackage{amsmath}
\usepackage{amsfonts}
\usepackage{graphicx}
\usepackage[noend,ruled,linesnumbered]{algorithm2e}
\usepackage{enumitem}
\usepackage{multirow}
\usepackage{booktabs}
\usepackage{subfigure}
\usepackage{bbm}

\usepackage{listings}
\usepackage{xcolor}

\newcommand{\mtop}{\textbf{MTOP}\xspace}
\newcommand{\eg}{\textit{e.g.}}

\usepackage{listings}
\usepackage{xcolor}

\colorlet{punct}{red!60!black}
\definecolor{background}{HTML}{EEEEEE}
\definecolor{delim}{RGB}{20,105,176}
\colorlet{numb}{magenta!60!black}

\lstdefinelanguage{json}{
    basicstyle=\normalfont\ttfamily,
    numbers=left,
    numberstyle=\scriptsize,
    stepnumber=1,
    numbersep=8pt,
    showstringspaces=false,
    breaklines=true,
    frame=lines,
    backgroundcolor=\color{background},
    literate=
     *{0}{{{\color{numb}0}}}{1}
      {1}{{{\color{numb}1}}}{1}
      {2}{{{\color{numb}2}}}{1}
      {3}{{{\color{numb}3}}}{1}
      {4}{{{\color{numb}4}}}{1}
      {5}{{{\color{numb}5}}}{1}
      {6}{{{\color{numb}6}}}{1}
      {7}{{{\color{numb}7}}}{1}
      {8}{{{\color{numb}8}}}{1}
      {9}{{{\color{numb}9}}}{1}
      {:}{{{\color{punct}{:}}}}{1}
      {,}{{{\color{punct}{,}}}}{1}
      {\{}{{{\color{delim}{\{}}}}{1}
      {\}}{{{\color{delim}{\}}}}}{1}
      {[}{{{\color{delim}{[}}}}{1}
      {]}{{{\color{delim}{]}}}}{1},
}

\usepackage[utf8]{inputenc}
\usepackage{array}
\usepackage{makecell}

\usepackage{inconsolata}

\title{All Birds with One Stone: Multi-task Text Classification for \\Efficient Inference with One Forward Pass}
\author{
Jiaxin Huang$^{1}$\footnotemark[1],
Tianqi Liu$^{2}$\footnotemark[1],
Jialu Liu$^{2}$,
Adam D. Lelkes$^{2}$,
Cong Yu$^{2}$,
Jiawei Han$^{1}$
\\ $^{1}$University of Illinois Urbana-Champaign ~~~ $^{2}$Google Research \\
{\texttt\small \{jiaxinh3, hanj\}@illinois.edu ~~~ }\\
{\texttt\small \{tianqiliu, jialu, lelkes, congyu\}@google.com}
}

\begin{document}

\maketitle
{
\renewcommand{\thefootnote}{\fnsymbol{footnote}}
\footnotetext[1]{Equal Contribution.\\ Corresponding author: \texttt{tianqiliu@google.com}}
}

\begin{abstract}
Multi-Task Learning (MTL) models have shown their robustness, effectiveness, and efficiency for transferring learned knowledge across tasks. In real industrial applications such as web content classification, multiple classification tasks are predicted from the same input text such as a web article. However, at the serving time, the existing multitask transformer models such as prompt or adaptor based approaches need to conduct $N$ forward passes for $N$ tasks with $O(N)$ computation cost. To tackle this problem, we propose a scalable method that can achieve stronger performance with close to $O(1)$ computation cost via only one forward pass. To illustrate real application usage, we release a multitask dataset on news topic and style classification. Our experiments show that our proposed method outperforms strong baselines on both the GLUE benchmark and our news dataset. Our code and dataset are publicly available at \url{https://bit.ly/mtop-code}.
\end{abstract}

\section{Introduction}
With recent progress in deep learning and natural language processing, pre-trained language models like BERT~\cite{Devlin2019BERTPO, Liu2019RoBERTaAR}, ELECTRA~\cite{Clark2020ELECTRAPT}, and TEAMS~\cite{shen2021training} exhibit strong transfer ability to downstream fine-tuning tasks such as classification. 
Apart from their exceptional performance on single applications, these models have also demonstrated their effectiveness as multi-task learners.

When it comes to industrial application such as classification for the web documents, there can be many labels or tasks to be predicted simultaneously with the same input format. 
For example, we may need to infer tens of labels of a blog post including its topic, sentiment, locality, safety, toxicity, etc. 

In the literature, multi-label classification has been proposed to tackle this problem where more than one label are assigned to each document at training time. 
But we argue that this multi-label formulation is sub-optimal for more realistic use cases. What we typically observe in practice is that the labels or tasks are developed gradually one after the other.  They often come with different class distributions, and vary in data sizes. More importantly, it is not guaranteed that these labels have overlap among training documents. In the extreme case, they can be disjoint. Being aware of this, we consider a more general multi-task learning (MTL) formulation where each document is associated with one task's label(s) at a time. Overlapped documents between tasks have multiple data records in the training corpus.

That being said, at prediction time, a model needs to simultaneously predict labels for all tasks. Most advanced MTL models built on top of Transformer are not efficient in this situation. The reason is that these models typically extract task-specific features at the bottom layers~\cite{Pilault2021ConditionallyAM} and this limits them from sharing intermediate representations. In the worst case, the time complexity can grow linearly with the size of tasks.
As the number of tasks can be large in real applications, it becomes infeasible to serve such models because of their long latency and high resource consumption. 
It becomes natural to seek for a unified multi-task model that trains on multiple tasks with little or no input overlap, and efficiently makes predictions for all tasks. 

In this paper, we aim to propose such a model that can infer labels for multiple text classification tasks in a scalable way without compromising quality.
Our approach is named \mtop, short for \textbf{M}ulti-\textbf{T}ask learning with \textbf{O}ne forward \textbf{P}ass at inference time. To the best of our knowledge, our work is the first to prioritize inference efficiency during the design of multi-task text classification. 
To achieve this goal, it becomes particularly challenging to ensure positive knowledge transfer across tasks. Many expensive but effective task-specific modules, such as conditional attention~\cite{Pilault2021ConditionallyAM}, cannot be used because of their cost. On the other hand, computationally cheaper modules, such as adding task-specific linear layers on top of a shared classification token~\cite{Liu2019MultiTaskDN}, turn out to be less helpful. 

By proposing a novel training recipe with well-tested task-specific modules, \mtop obtains new state-of-the-art results on news and GLUE datasets, and also the best
efficiency in serving. 
We summarize our contributions as follows:
\begin{itemize}[leftmargin=*]
\item We design a new scalable multi-task architecture that adapts pretrained weights with only one forward pass at the inference stage through trainable prompts and a conditional pooler layer. The proposed architecture is scalable to a large number of tasks and reduces the negative transfer among tasks.
\item We propose an effective training recipe to mitigate task conflicts. A better initialization is used so that knowledge can be transferred from single task pre-finetuning to the multi-task model. To avoid gradient conflict among tasks during fine-tuning, a simple yet effective stop-gradient mechanism is applied over learnable prompts.
\item To illustrate the real-world application of our approach, we release a multi-task news classification dataset with eight tasks, where training documents are owned by each task and evaluation documents are shared among tasks.
\end{itemize}

\section{Related Work}
Multi-task Learning (MTL), which was inspired by the ability of humans to transfer knowledge from previously mastered activities to new ones, has been applied to a wide range of tasks beyond the field of NLP~\cite{caruana1997multitask, Ruder2017AnOO, Donahue2014DeCAFAD}. 
Recent studies in NLP also make successful efforts to train multi-task models to outperform independently fine-tuned ones on the benchmark datasets. They are focused on developing advanced techniques in four directions. The first direction is to explore task transferability via cross-task learning. The second direction is to improve the model architectures. The third direction is to distill a general multi-task student from single task teachers. The last direction is to study the effective training dynamics.

\paragraph{Cross-task Learning}
By multi-task pre-training on a mixture of language understanding generation tasks, 
ExT5~\cite{aribandi2021ext5} and FLAN~\cite{wei2021finetuned} effectively transfer knowledge between tasks and substantially surpass single-task fine-tuning. Meta-learning~\cite{tripuraneni2021provable, finn2017model} learns about one's own learning and learning processes. With that, the model can easily adapt to new tasks. Researchers also study how to effectively do few-shot learning for a given task from existing learned tasks~\cite{wang2020generalizing}.

\paragraph{Model Architectures}
As an influential early work exploring multi-task BERT models, MT-DNN~\cite{Liu2019MultiTaskDN} attaches task-specific linear layers on top of a shared Transformer backbone. Since the backbone is shared, it may introduce negative transfer among tasks. With the growing size of language models, light-weight tuning methods have also been proposed to reduce the number of parameters needed for training a multi-task model. To tackle the above challenges, prompt-based~\cite{lester2021power} and adapter-based~\cite{Pilault2021ConditionallyAM} models are proposed.

Prompts are typically task descriptions or examples prepended to the input of language models. For example, GPT-3~\cite{Brown2020LanguageMA} takes manually designed prompts as guidance to generate answers for specific tasks. Prompt engineering~\cite{Jiang2020HowCW,Schick2021ExploitingCF,Shin2020ElicitingKF} has been studied to search for high-quality templates, either manually or automatically. However, the non-differentiable nature of discrete tokens can lead to sub-optimal results, which motivated the introduction of \emph{soft prompts}, where continuous vectors replace discrete tokens. Prompt tuning~\cite{Lester2021ThePO} prepends $k$ tunable soft prompts to input tokens per downstream task. Prefix-tuning~\cite{Li2021PrefixTuningOC} prepends continuous task-specific vectors to each layer in the encoder. 

Adapters are small trainable modules parameterized by task-specific embeddings. These modules are inserted into, or used to replace, key modules to make the Transformer model adaptive to new tasks, so that it can project the original rich information to a task-related space. 
\cite{Houlsby2019ParameterEfficientTL} adds a feed-forward bottleneck and fine-tunes the layernorm of each Transformer layer.
CA-MTL~\cite{Pilault2021ConditionallyAM} remodulates the self-attention layer, normalization layer, embedding projection layer, and bottleneck layer by conditional weight transformation specified by task-specific vectors. By integrating local task modules to the global task agnostic module,
\citet{Stickland2019BERTAP}, \citet{mahabadi2021compacter}, and \citet{Tay2021HyperGridTT} mix efficient adapters to BERT or T5~\cite{Raffel2020ExploringTL}. 

\paragraph{Distillation}
BAM~\cite{Clark2019BAMBM} learns a multi-task model by distilling knowledge from single-task teacher models. \citet{wei2021flexible} explored partial fine-tuning task specific layers through distillation so that the computation time during finetuning is reduced. Although their method makes predictions for multiple tasks in one forward pass, it is not scalable since its serving latency deteriorates significantly as the number of tasks grows.

\paragraph{Training Dynamics}
Optimization for MTL is broadly studied to mitigate the issues via gradient modulation and task scheduling. 
Traditional approaches sample tasks uniformly~\cite{caruana1997multitask} or proportionally to data size~\cite{Sanh2019AHM}. Subsequent studies~\cite{Gottumukkala2020DynamicSS, Pilault2021ConditionallyAM, Sharma2017OnlineML} have explored dynamic sampling strategies based on the current performance of the model. Another line of work~\cite{Yu2020GradientSF, Wang2021GradientVI} focuses on eliminating conflicting gradients from multiple tasks. 

\section{Our Multi-Task Setting}
In this work, we mainly focus on text classification tasks.
Take news classification as an example, where each news article is expected to be classified into one or more topics (\eg, politics, technology, business) and styles (\eg, opinion, interview, live blog). 
Different datasets are usually collected using distinct crowdsourcing instructions. This is because (1) the industry or company can keep improving its news eco-system by defining new sets of tasks, and more importantly, (2) it is inefficient to collect labels of all tasks at the same time, since different topics or styles may have low overlaps among training documents; using the same documents for all labels would cause label imbalance. Thus the training data typically consists of a set of independent datasets. At the evaluation stage, the model can be evaluated on disjoint datasets or a joint dataset with partial or all task labels. 
However, at  serving time, all task labels of an unseen document need to be generated.


\section{Methodology}

\begin{figure*}[t]
\centering
\centering
\includegraphics[width=1.0\textwidth]{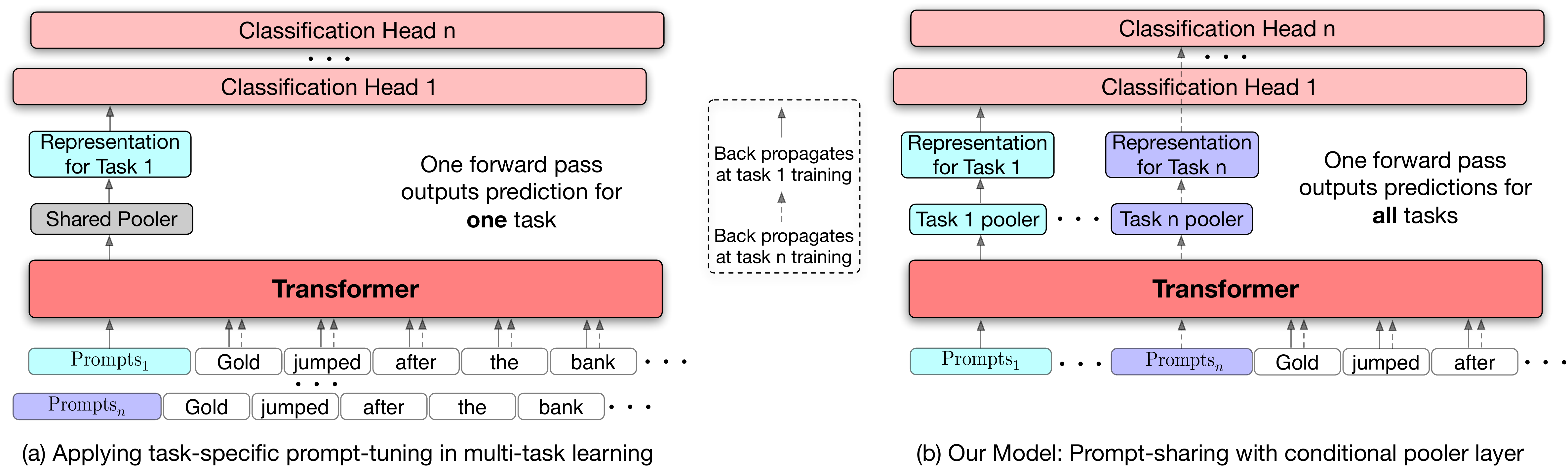}
\caption{Comparison between task-specific prompt-tuning and \mtop in multi-task learning. (a) Task-specific prompts are prepended separately to the input, and at inference stage the Transformer outputs prediction for one task in one forward pass. (b) Our model \mtop allows for outputing prediction for all tasks in one forward pass. \mtop concatenates prompts for multiple tasks and prepends them to the input text, with gradient backpropagation only applied to prompts of the task being trained to reduce negative transfer among tasks. Meanwhile, we apply conditional pooler layer to better align the hidden output presentation with task-specific label space.
}
\label{fig:share_prompt}
\end{figure*}

To preserve inference efficiency and positive transfer among tasks at the same time, we design a scalable model that is able to capture task-specific features at the bottom layers and use only one forward pass to infer labels for all tasks.
We make three adaptations on the Transformer model architecture 
and training approaches: 
(1) To capture features for all tasks at the bottom layers, we concatenate and prepend multiple task-specific prompts before the input sequence, and
apply a gradient stopping policy to avoid conflict of gradients.~(2)~We propose a conditional pooler layer 
to induce different latent space of tasks and reduce negative transfer.~(3) We improve the training procedure by carefully designing a simple yet effective initialization approach for the prompts and the conditional pooler. 

We will introduce the details of our method in the following sections. Specifically, we will first cover preliminaries of prompt tuning in Sec.~\ref{sec:prompt_prelim}, and then introduce how we approach the multi-task challenge in Sec.~\ref{sec:prompt_share} and Sec.~\ref{sec:init}. Finally, we will discuss about the scalability of our method in Sec.~\ref{sec:scale}.

\subsection{Preliminaries on Prompt Tuning}\label{sec:prompt_prelim}
Consider we have a pre-trained Transformer model with pre-trained weights $\phi$. A traditional classification model takes a sequence of tokens $X$ with length $L_X$ as input and takes the output of the Transformer at the \texttt{[CLS]} token followed by a $\tanh$-pooler layer and a classification head to calculate the class probability $\mathbb{P}_\phi(Y|X)$, with $Y$ being the ground truth label of $X$.
In prompt-based methods, a sequence of task-specific prompt tokens $P$ of length $L_P$ are prepended to the input $X$. The prompt tokens $P$ in GPT-3~\cite{Brown2020LanguageMA} zero-shot and few-shot learning are manually selected from a fixed vocabulary with fixed embeddings. In prompt tuning~\cite{Lester2021ThePO}, the soft prompt tokens are a new set of parameters $P^e \in \mathbb{R}^{L_P\times d}$ which can be updated during training, with $d$ being the dimension of word embeddings. Thus the training objective is to maximize $\mathbb{P}_\phi(Y|[P^e;X^e])$, where $X^e$ is the embedding matrix of the input sequence $X$. The pre-trained weights $\phi$ are kept fixed because prompt-tuning seeks for efficient and tractable finetuning for large language models. In our case, our goal is to train a serving-efficient multitask model with small to moderate size, thus we enable updating $\phi$ to optimize performance and encourage positive transfer among tasks.

\subsection{Our Adaptations to Multi-task Learning}\label{sec:prompt_share}

\paragraph{Concatenating Prompts with Gradient Stopping Policy.}
A simple way to fine-tune $N$ multiple tasks with prompts is pictured in Fig.~\ref{fig:share_prompt}(a). A prompt tensor $P^e_{1:N} \in \mathbb{R}^{N\times L_P\times d}$ is initialized, and when task $i$ is being trained, the model slice through the first dimension of $P^e_{1:N}$ to get $P^e_i \in \mathbb{R}^{L_P\times d}$ (corresponding to the prompts of task $i$) and prepend it to the input text.
During inference, in order to make predictions for all $N$ tasks, we need to prepend prompts for each task, resulting in $N$ forward passes. One direct way of reducing this computation cost is to concatenate prompts for different tasks to get a shared prompt pool $\{P^e_1, \dots, P^e_N\} \in \mathbb{R}^{(NL_P)\times d}$, which is later prepended to the input sequences, as shown in Fig.~\ref{fig:share_prompt}(b). 

However, the above direct way of concatenating prompts does not lead to good performance, which we will show in our experiments in Sec.~\ref{sec:ablation}. This is due to possible negative transfer between tasks. For example, when training a batch of data in task $i$, backpropagation not only updates the corresponding prompt embedding $P^e_i$, but also affects other prompt embeddings $P^e_j (j\neq i, 1\leq j\leq N)$.

Given the above concern, we only allow the gradients of task $i$ to backpropagate to prompt embedding $P^e_i$, and stop the gradients for other prompts, as shown in Fig.~\ref{fig:share_prompt}(b). Prompt sharing with gradient stopping benefits the performance of the model in two aspects: (1) Multiple prompts are input together into the self-attention modules so related tasks are easy for the Transformer model to identify and refer to; and (2) non-related tasks will less likely impair the results of each other due to the gradient stopping policy.

\paragraph{Conditional Pooler Layer.}
On the output side, the \texttt{[CLS]} token is no longer used to represent the whole sentence for classification, because it is shared among the tasks. We use an average pooling layer over the encoder outputs of the corresponding task prompts to obtain task-specific sentence representations. We denote the overall encoder outputs of the last layer of the Transformer model as $H\in\mathbb{R}^{(NL_P+L_X)\times d}$, and $E_i$ is the sentence representation for task $i$:
\begin{equation}\label{eq:avg_hidden}
E_i = \frac{1}{L_P} \sum_{j=(i-1)L_P+1}^{iL_P}{H_j}
\end{equation}

In standard Transformer models like BERT, a pooler layer parameterized by $\{W, b\}$ projects the encoder output to task-specific classification layers. In multi-task training, previous studies commonly share the pooler layer among all tasks before loading to classification layer $V_i$ (see Fig.~\ref{fig:share_prompt}(a)), and the final prediction is derived by
\begin{equation}\label{eq:prompt_output}
\mathbb{P}_\phi(Y|X) = \text{Softmax}\left(V_i \, \tanh \left( W E_i + b \right) \right)
\end{equation}
However, since the sentence representations of all tasks will go through the same kernel $W$ and share the same column space, and the pooler layer is right before the classification layer, this may introduce strong task interference.
Therefore, we propose task-dependent pooler layers as shown in Fig.~\ref{fig:share_prompt}(b) and derive the prediction probability for task $i$ as
\begin{equation}\label{eq:prompt_output}
\mathbb{P}_\phi(Y|X, i) = \text{Softmax}\left(V_i \, \tanh \left( W_i E_i + b_i \right) \right)
\end{equation}
where $\{W_i\in\mathbb{R}^{d\times d}, b_i\in\mathbb{R}^d\}$ parameterize the conditional pooler layer and
$V_i\in\mathbb{R}^{d\times c_i}$ is the classification head, with $c_i$ being the number of classes for task $i$.

\subsection{Initialization of Prompts and Conditional Poolers}\label{sec:init}
Previous studies~\citep{Clark2019BAMBM, ghiasi2021multi} have explored effective knowledge transfer from single-task experts to multitask models. Their general philosophy is to first train several single task models to generate pseudo labels, and then distill the knowledge by using the pseudo labels to train a new multitask student model. 
Distillation requires training single task teacher models, generating pseudo labels, and modifying the loss functions, which introduce additional complexity to the system. To simplify the process, in our system, we partially train single task models and initialize additional task-specific parameters (prompts and conditional pooler layers) from them.


We first freeze the pretrained backbone and train single task models with $L_P$ prompts prepended to the input. For each task, we only train the prompt embeddings $P^e_i$, the pooler layer ($W_i$, $b_i$), and the classification head $V_i$. 
Then we use these learned parameters to initialize the prompt embeddings and conditional pooler layers in our multi-task model \mtop.
There are two advantages for our initialization approach:
(1) we keep the single-task training phase more efficient by freezing the parameters in the embedding table and transformer layers; (2) \mtop benefits from the initialized prompts and conditional pooler layers with better task-related knowledge extracting ability at both the input and output side of the model, leading to a smoother training of \mtop.

\subsection{Scalability}\label{sec:scale}
In this section, we discuss the scalability of \mtop with respect to the number of tasks.
Unlike adding task-specific transformer layers~\citep{wei2021flexible}, our approach is scalable in increasing number of tasks. Each additional task with $c$ classes  will only introduce $L_P$ prompts with $L_P\cdot d$ parameters, a conditional pooler with $d^2+d$ parameters, and a classification head with $d\cdot c$ parameters. 
So the total number of model parameters a new task introduces is $d^2 + (L_p+c+1)d$. 

For instance, for the eight-task GLUE benchmark, \mtop only has around $4\%$ more parameters than a single-task model we set $L_p=2$ in our experiments).



\section{News Headline Classification (NHC) Dataset}
In a real-world system such as a news publisher or aggregator, a machine learning system should label any given news article with a pre-defined set of topic and style labels. These labels are used to organize the news articles in different hubs or sections. The datasets of different tasks can be collected at different stages, using different crowdsourcing instructions. To illustrate this real-world application scenario, we collect a multitask news dataset named as News Headline Classification (NHC) Dataset.

We derive our dataset from a Kaggle news category dataset\footnote{\url{https://www.kaggle.com/datasets/rmisra/news-category-dataset}}~\cite{misra2018news,misra2021sculpting}. The original dataset contains around 200k news headlines from the year 2012 to 2018 obtained from HuffPost. An example of the original data is:
\begin{lstlisting} [basicstyle=\ttfamily\small,breaklines]
{
  "category": "CRIME",
  "headline": "There Were 2 Mass Shootings In Texas Last Week, But Only 1 On TV",
  "authors": "Melissa Jeltsen",
  "link": "https://www.huffingtonpost.com/entry/texas-amanda-painter-mass-shooting_us_5b081ab4e4b0802d69caad89",
  "short_description": "She left her husband. He killed their children. Just another day in America.",
  "date": "2018-05-26"
}
\end{lstlisting}

The original dataset has 40+ news topics with a skewed distribution, which is a common phenomenon in real-world problems. When developing classifiers for each task, practitioners often collect a balanced dataset with sample size no less than certain number. To illustrate this scenario, we first selected the top 7 most popular categories to create a binary multi-task classification dataset. Specifically, we sampled 2k examples from each category as the positive ones, and then randomly sampled 2k examples that are not marked as the target category as the negative ones, so in total we have $4k\times 7 = 28k$ examples collected. 
For each task we designate a randomly sampled $80\%$ as the training set and the remaining $20\%$ as the evaluation set. 
To illustrate the real-world use case,
we further collect style labels of whether the news article is opinionated or not on nearly $3k$ randomly sampled examples and $800$ randomly sampled evaluation examples. We use a proprietary crowdsourcing platform to label the data. 


\section{Experiments and Results}
We use two datasets to evaluate our proposed method \mtop and compare with other strong baselines. We first evaluate on the NHC dataset to demonstrate the model's performance in a realistic industrial news classification scenario. The second dataset is the standard GLUE benchmark~\citep{Wang2018GLUEAM} to demonstrate that our model can be applied to more general classification tasks.
Furthermore, we provide a series of module analyses to evaluate the performance of each module.

\subsection{Experiment Setup}\label{sec:exp_setup}
\paragraph{Backbone Model Selection} We use the best BERT base model as our backbone, which is the ELECTRA-base++~\cite{Clark2020ELECTRAPT}\footnote{There are two base ELECTRA models trained from different datasets. Here we use the base++ version derived from the XLNET dataset.} as the backbone model for our method and baselines. ELECTRA-base++ has the identical network architecture as BERT base. We set the maximum sequence length to be $128$. 

\paragraph{Hyperparameters} We choose the best hyperparameters based on the model's performance on the GLUE dev set: for prompt sharing, we use $2$ task-specific prompt tokens per task in GLUE. For training, we use a batch size of $16$ for each task, and let the model train for $5$ epochs for GLUE and $20$ epochs for NHC. We set the learning rate to be $1\mathrm{e}{-5}$ and use linear warm-up schedule with 10\% warm-up steps.  We use a proportional sampling strategy to sample one task for each batch with weight proportionate to the task size.

\paragraph{Checkpoint Picking} For each model and dataset, we repeat training five times. For each repetition, we use the best checkpoint with the highest average evaluation metric (this checkpoint is shared among all tasks). Note that this way we enable one model to be used in serving time, unlike other MTL studies which report the best evaluation metric for each task from different checkpoints. After we get the evaluation results for each run, we report the median over 5 runs. For the GLUE test score, we choose the median checkpoint and submit its predictions to the GLUE leaderboard.

\paragraph{Initialization Approach} 
We initialize the backbone from the pretrained ELECTRA-base++. 

To prepare for the initialization of prompts and conditional pooler layers, we fine-tune single-task models with $2$ prompts prepended to the input sequence. We only train the prompts, pooler layer, and classification head, while keeping the backbone frozen. We denote these trained models as $\{M_i\}_{i=1}^N$ where $i$ is the task index. We describe several of our model variants as follows.

We experiment with three choices for prompt initialization:
\vspace{-0.5em}
\begin{itemize}[leftmargin=*]
\setlength\itemsep{-0.4em}
\item Random initialization (RD). 
For the $i$th prompt, we draw a $768$-dimensional random vector $P^e_i$ with each element $P^e_{ij}\stackrel{iid}{\sim} \text{TruncatedNormal}(\mu=0, \sigma=0.02)$, meaning that we discard and re-draw any samples that are more than two standard deviations ($0.04$) from the mean.

\item Token initialization (TK). Following~\cite{Lester2021ThePO}, we first construct a common vocabulary file with the most popular $5k$ tokens from the corresponding dataset. Then we initialize each prompt by the embedding of a token randomly sampled from the vocabulary file.

\item Initialization from single tasks (ST). We initialize task-specific prompts with weights trained from single tasks as described in Sec.~\ref{sec:init}.
\end{itemize}
\vspace{-0.5em}
For the conditional pooler layer, we have two choices of initialization.
\vspace{-0.5em}
\begin{itemize}[leftmargin=*]
\setlength\itemsep{-0.5em}
\item Random initialization (RD). The kernels are initialized from GlorotUniform distribution~\cite{Glorot2010UnderstandingTD} and biases are initialized as zero.
\item Initialization from single tasks (ST). We initialize conditional pooler layers with weights trained from single tasks as described in Sec.~\ref{sec:init}.
\end{itemize}


\subsection{Baseline Methods}
We compare with several important multi-task learning baselines and recently proposed prompt-tuning methods, as well as with single-task fine-tuning methods. For all baselines, we load ELECTRA-base model as the encoder, in order to control the backbone and only evaluate the effect of the MTL approach. The compared methods are listed below:
\begin{itemize}[leftmargin=*]
\item \textbf{Single Task}: Single task fine-tuning on each GLUE or NHC task independently. 
For each task, we choose the best checkpoint according to the evaluation metrics.
\item \textbf{MT-DNN}~\cite{Liu2019MultiTaskDN}: A multi-task deep neural network that uses the Transformer encoder as a shared architecture among tasks, and the top layer is attached to task-specific linear layers. 
Only one checkpoint is picked for evaluation, which is a fair comparison to our approach. This could result in lower evaluation metrics compared to using some standard tricks (such as training MNLI first for some other tasks) and using different best checkpoints for different tasks.

\item \textbf{P-Tuning}~\cite{Lester2021ThePO}: Prompt tuning method that trains task-specific prompts on each task. The prompts are initialized from random common tokens and we make the backbone trainable as well for better performance.

\item \textbf{P-Tuning v2}~\cite{Liu2021PTuningVP}: A recently proposed advanced deep prompt tuning method that adds independent prefix tokens in each transformer layer. Again, the prompts are initialized from random common tokens and we make the backbone trainable as well.

\item \textbf{CA-MTL}~\cite{Pilault2021ConditionallyAM}: A recently proposed multi-task model that uses adaptive modules parameterized by task embeddings.
\end{itemize}

\subsection{Evaluation on the NHC Dataset}
We evaluate our proposed model \mtop and other baseline methods on NHC evaluation dataset of 8 tasks in Table~\ref{tab:nhc_results}. As shown in the table, all MTL methods except MT-DNN have better average score than the single task fine-tuning method, indicating the effectiveness of sharing the same backbone architecture between tasks with flexible modules. On the other hand, MT-DNN is the most efficient one among the baseline approaches with only one forward pass to compute all predictions for all tasks.
Among all the methods, \mtop achieves the best average score with $1.4$ points improvement over single task models. This indicates that effectively transferring knowledge among tasks helps the model generalize well among multiple tasks. P-Tuning v2 does not work well when we enable the training on the backbone compared with P-Tuning. We notice that P-Tuning also performs well on the NHC dataset. However, it requires $8$ forward passes for one batch of examples during serving, while \mtop only requires one. Notably, \mtop can achieve both a performance boost and a serving efficiency improvement with careful design of multitask modules and training recipes.

\setlength{\tabcolsep}{4.2pt}
\begin{table*}[t]
\small
\centering
\begin{tabular}{c|c|cccccccc|c}
\toprule
\thead{Methods}  & \thead{\# Forward \\ Pass} & \thead{Politics} & \thead{Wellness} & \thead{Enter- \\ tainment} & \thead{Travel} & \thead{Style and \\ Beauty} & \thead{Parent-\\ing} & \thead{Healthy \\Living} & \thead{Opinion} & \thead{Avg}\\
\midrule
\textbf{Single Task} & 8 & 88.1 & 86.4 & 90.1 & 93.5 & 92.4 & 87.1  & 82.6 & 76.0 & 87.0\\
\textbf{MT-DNN} & 1 & 88.0 & 86.2 & 89.7 & 93.4 & 90.7 & 87.9 & 84.1 & 74.9 & 86.9\\
\textbf{P-Tuning} & 8 & 88.6 & 87.3 & 90.2 & 94.2 & 93.1 & 89.4 & \textbf{85.5} & \textbf{76.4} & 88.1\\
\textbf{P-Tuning v2} & 8 & 88.1 & 87.1 & 90.7 & 92.7 & 91.1 & 88.2 & 84.1 & 74.1 & 87.0\\
\textbf{CA-MTL} & 8 & 88.3 & 87.6 & 90.4 & 94.1 & 93.2 & \textbf{89.9} & 82.7 & 74.6 & 87.6 \\
\midrule
\mtop & 1 & \textbf{89.9} & \textbf{88.2} & \textbf{91.6} & \textbf{94.6} & \textbf{93.3} & 88.6 & 85.2 & 75.5 & \textbf{88.4}\\
\bottomrule
\end{tabular}
\vspace{-0.5em}
\caption{
Accuracy of all methods on the evaluation set of NHC dataset. We repeat each experiment $5$ times and report the median.}
\label{tab:nhc_results}
\end{table*}
\setlength{\tabcolsep}{6pt}

\subsection{Evaluation on GLUE}
We also evaluate \mtop and other baseline methods on the GLUE test set of 8 tasks in Table~\ref{tab:glue_results}. 
As shown in the table, the multi-task learning methods have higher average score than the single task fine-tuning method, indicating the effectiveness of sharing the same backbone architecture between tasks. The task benefiting the most from MTL is RTE, because it is a natural language inference task and can benefit from effective knowledge transfer from MNLI and MRPC. Interestingly, single task models perform well on tasks with large training data such as MNLI, QNLI, QQP, and SST-2. The gap between \mtop and single task models is much smaller comparing to other MTL variants, which indicates that negative transfer is effectively reduced. On the other hand, for small tasks such as MRPC, RTE, and STS-B, \mtop shows superior performance, which demonstrates the effective positive transfer among tasks with flexible module design and initialization. CoLA does not show improvement from single task to multitask. We believe this is because CoLA, as a grammatically acceptance task, is less related with the rest of the semantic-based tasks, and benefits the least from multi-task learning. Among all the methods, \mtop achieves the best average score and also the best serving efficiency. Notably, \mtop outperforms single-task models by $1.5$ points.

\setlength{\tabcolsep}{4.2pt}
\begin{table*}[t]
\small
\centering
\begin{tabular}{c|c|cccccccc|c}
\toprule
\thead{Methods}  & \thead{\# Forward \\ Pass} & \thead{CoLA \\(8.5k)} & \thead{MNLI\\(392.7k)} & \thead{MRPC\\(3.7k)} & \thead{QNLI\\(104.7k)} & \thead{QQP\\(363.8k)} & \thead{RTE\\(2.5k)} & \thead{SST-2\\(67.3k)} & \thead{STS-B\\(5.7k)} & \thead{Avg}\\
\midrule
\textbf{Single Task} & 8 & \textbf{63.3} & \textbf{88.4}/\textbf{87.7} & 88.9 & \textbf{93.1} & \textbf{72.6}  & 67.0 & 95.2 & 88.9 & 82.8\\
\textbf{MT-DNN} & 1 & 60.6 & 87.6/87.1 & 87.7 & 92.9 & 72.2 & 80.2 & 94.8 & 88.6 & 83.5\\
\textbf{P-Tuning} & 8 & 60.3 & 87.9/87.3 & 88.4 & 92.8 & 71.8 & 80.4 & 95.5 & 89.4 & 83.8\\
\textbf{P-Tuning v2} & 8 & 61.8 & 88.0/87.4 & 88.0 & 92.7 & 72.2 & 80.5 & 95.3 & 88.8 & 83.9\\
\textbf{CA-MTL} & 8 & 61.5 & 87.9/87.4 & 88.3 & 92.6 & 72.2 & 80.4 & 95.3 & 88.7 & 83.8 \\
\midrule
\mtop & 1 & 62.3 & 88.1/87.6 & \textbf{89.3} & 92.6 & 72.4 & \textbf{80.6} & \textbf{95.6} & \textbf{90.0} & \textbf{84.3}\\
\bottomrule
\end{tabular}
\vspace{-0.5em}
\caption{
Comparison among all methods on GLUE test set. We report Matthew's correlation on CoLA, Spearman's correlations on STS-B, F1 on QQP/MRPC, and accuracy for other GLUE tasks. For MNLI, the first number is MNLI match and the second is MNLI mismatch. We repeat each experiment $5$ times and use the median checkpoint from the dev set to submit to GLUE leaderboard. The numbers in parentheses under the task names are the training sample sizes.}
\label{tab:glue_results}
\end{table*}
\setlength{\tabcolsep}{6pt}

\subsection{Ablation Study}\label{sec:ablation}
In this section we analyze the effectiveness and robustness of each modification we proposed in \mtop via ablation studies.

\paragraph{Length of Prompts}
When introducing prompts, there are two questions to be asked. The first is whether we need additional task-agnostic prompts shared among tasks for larger model capacity. The second is the optimal length of task-agnostic prompts and task-specific prompts.

Task-agnostic prompts are prepended before the concatenated task-specific prompts and randomly initialized from a truncated normal distribution with standard deviation $0.02$. Their gradients are never stopped and they are not used in computing the task-specific representation in Equation~\ref{eq:avg_hidden}. 

\begin{figure}[t]
\centering
\includegraphics[width=0.9\linewidth]{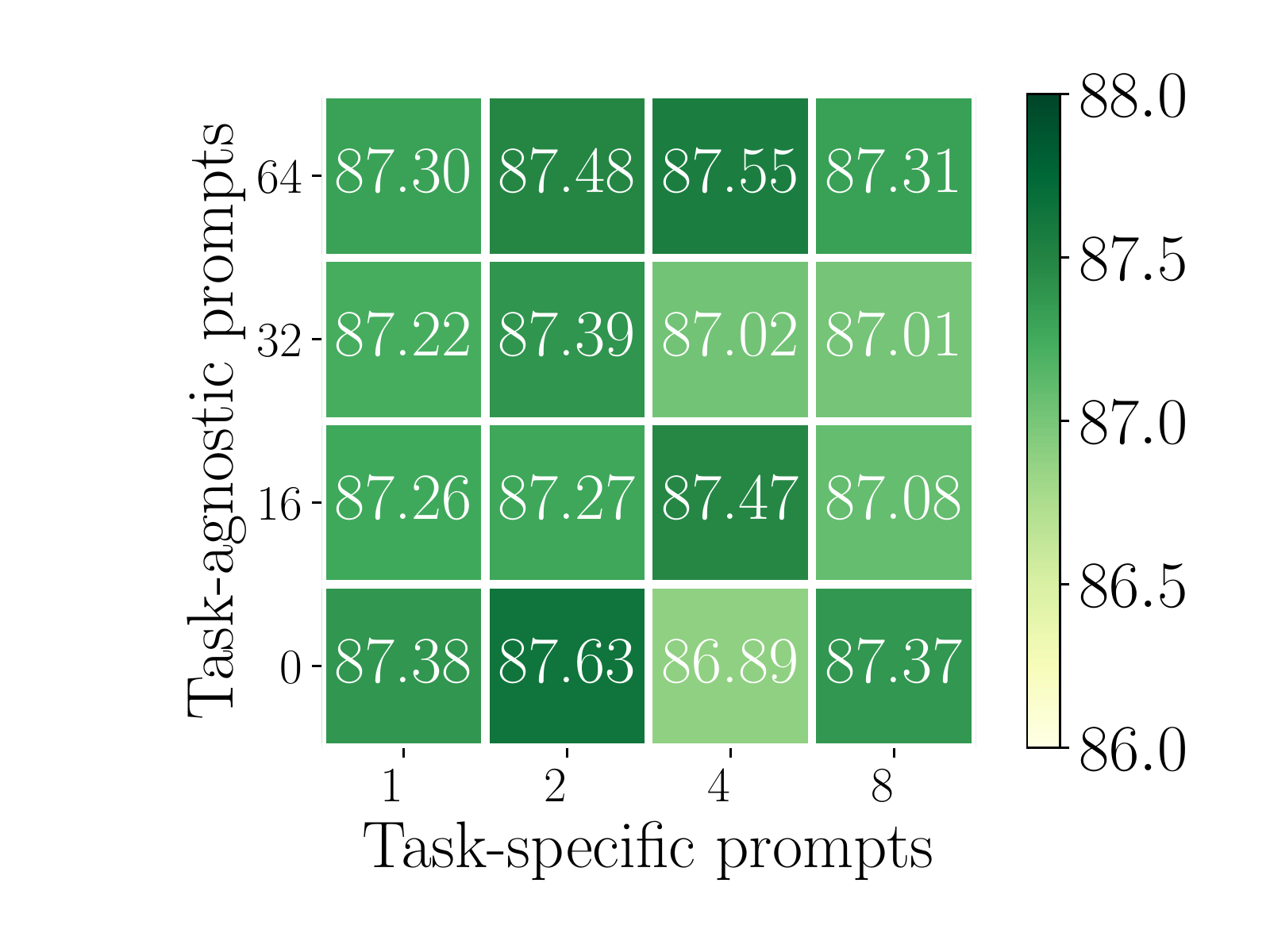}
\caption{Performance of model on GLUE dev set with different numbers of task-specific prompt tokens per task and shared task-agnostic tokens.}
\label{fig:length_ablation}
\end{figure}

As shown in Figure~\ref{fig:length_ablation}, we found that zero shared task-agnostic prompts and two task-specific prompts work the best on GLUE dev set.

\setlength{\tabcolsep}{4.2pt}
\begin{table*}[t]
\small
\centering
\begin{tabular}{c|cccccccc|c}
\toprule
\thead{Methods} & \thead{CoLA} & \thead{MNLI} & \thead{MRPC} & \thead{QNLI} & \thead{QQP} & \thead{RTE} & \thead{SST-2} & \thead{STS-B} & \thead{Avg}\\
\midrule
\textbf{(1) MT-DNN} & 62.9 & 87.7/87.2 & 85.5 & 92.1 & 91.2 & 83.7 & 94.3 & 90.6 & 86.0\\
\textbf{(2) Prompts(RD) +\ Cond(RD)} & 63.3 & 87.7/87.3 & 88.5 & 92.5 & \textbf{91.5} & 84.5 & 94.5 & 91.0 & 86.8\\
\textbf{(3) Prompts(TK) +\ Cond(RD)} & 64.4 & \textbf{87.8}/87.2 & 89.2 & 92.1 & 91.4 & 84.8 & 94.0 & 91.2 & 86.9\\
\textbf{(4) Prompts(ST) +\ Cond(RD)} & 65.9 & 87.7/87.2 & 90.0 & \textbf{92.6} & 91.3 & 85.9 & 94.6 & 91.6 & 87.4\\
\textbf{(5) Prompts(RD) +\ Cond(ST)} & 65.3 & \textbf{87.8}/87.1 & 89.2 & 92.3 & 91.4 & 85.6 & 94.3 & 91.5 & 87.2\\
\textbf{(6) \mtop\ w/o SG} & 66.9 & 87.7/\textbf{87.5} & \textbf{90.2} & 92.4 & 91.4 & 84.8 & 94.4 & \textbf{91.8} & 87.5 \\
\textbf{(7) \mtop} & \textbf{67.2} & \textbf{87.8}/\textbf{87.5} & 89.0 & 92.4 & 91.4 & \textbf{86.6} & \textbf{94.7} & 91.6 & \textbf{87.6}\\
\bottomrule
\end{tabular}
\vspace{-0.5em}
\caption{
Ablation studies on GLUE dev dataset. We report Matthew's correlation on CoLA, average of Pearson's and Spearman's correlations on STS-B, and accuracy for other GLUE tasks. For MNLI, the first number is MNLI match and the second is MNLI mismatch. We repeat each experiment $5$ times and report the median.}
\label{tab:ablation}
\end{table*}
\setlength{\tabcolsep}{6pt}

\paragraph{Module Analysis}
This section studies the effect of several proposed modules and training methods in \mtop: prompts and conditional poolers, as well as their initialization and the gradient stopping policy. We study the following ablations:

\vspace{-0.5em}
\begin{enumerate}[leftmargin=*,label={(\arabic*)}]
\setlength\itemsep{-0.4em}
\item MT-DNN: no prompts are used and all tasks share the same pooler layers.
\item Prompts(RD) +\ Cond(RD): our proposed concatenated task prompts and conditional layers are applied to the model architecture, but are randomly initialized as described in Section \ref{sec:exp_setup}. We stop the gradient (SG) on prompts of other tasks when training the current one.
\item Prompts(TK) +\ Cond(RD): the same as (2), except that we initialize the prompts by common tokens as described in Section \ref{sec:exp_setup}.
\item Prompts(ST) +\ Cond(RD): the same as (2), except that we initialize the prompts by single task trained weights as described in Section \ref{sec:exp_setup}.
\item Prompts(RD) +\ Cond(ST): the same as (2), except that we initialize the conditional poolers by single task trained weights as described in Section \ref{sec:exp_setup}.
\item \mtop\ w/o SG: The same as \mtop (prompts and conditional poolers are both initialized from single task trained weights) except that we allow gradient flow to other prompts while training the current task.
\end{enumerate}
\vspace{-0.5em}

We show the performance of all ablations in Table \ref{tab:ablation}. By comparing (1) with (2), we show that adding randomly initialized prompts and conditional poolers can already boost the performance of various tasks, especially on CoLA, MRPC, and STS-B. By comparing (2), (3) and (4), we show that using ST to initialize prompt embeddings is superior to RD and TK. By comparing (2) and (5), we also demonstrate that it is essential to initialize conditional pooler layers with single-task trained weights.
The comparison between (5) and (6) shows that by stopping gradient and only updating prompts of the task being trained, the model is more robust on small tasks such as CoLA and RTE. This is because large tasks can occupy resources from other task-specific prompts.

\section{Conclusion}
In this paper we propose an inference-efficient multi-task learning algorithm that learns a generalized representation and can output predictions for all tasks in one forward pass without a quality compromise. We propose prompt sharing with a gradient stopping policy and conditional pooler layers, as well as an effective initialization scheme for the introduced parameters. Our method achieves strong performance and outperforms previous state-of-the-art on the GLUE benchmark dataset as well as our NHC news dataset, which we are releasing publicly.

Several interesting directions can be explored in future studies. For our prompt module, we currently add two prompts before the input sequence. This may still face scalability issue with thousands of tasks, in which case we can merge prompts by task groups via sharing prompts or mixture-of-experts layers.
We can also explore a dynamic scheduling module for effective task selection. Moreover, since our architecture is focused on document-level tasks, it would be interesting to extend the study to token-level tasks where each token has a set of labels to be predicted.

\bibliography{ref}
\bibliographystyle{acl_natbib}

\end{document}